\documentclass[conference]{IEEEtran}
\usepackage{times}

\usepackage[numbers]{natbib}
\usepackage{multicol}
\usepackage[bookmarks=true]{hyperref}
\usepackage{balance}
\let\labelindent\relax
\usepackage{enumitem}
\usepackage{graphicx}
\usepackage{amsmath}
\usepackage{subcaption}
\usepackage{multirow}
\usepackage{multicol}
\usepackage{xcolor}
\usepackage{soul}
\usepackage{fdsymbol}
\definecolor{purple}{RGB}{210, 0, 210} 
\definecolor{green}{RGB}{0, 153, 0} 
\definecolor{grey1}{RGB}{255, 255, 255}
\definecolor{grey2}{RGB}{160, 160, 160}
\definecolor{grey3}{RGB}{0, 0, 0}

\usepackage{amsfonts}

\begin{document}

\title{DULA: A Differentiable Ergonomics Model for Postural Optimization in Physical HRI}


\author{\authorblockN{Amir Yazdani$^*$, Roya Sabbagh Novin$^*$, Andrew Merryweather$^*$, Tucker Hermans$^{*\dagger}$}
\authorblockA{$^*$University of Utah Robotics Center, Salt Lake City, UT, $^\dagger$NVIDIA, Seattle, WA\\
Email: amir.yazdani@utah.edu}
}


%

\maketitle

\begin{abstract}
Ergonomics and human comfort are essential concerns in physical human-robot interaction applications. Defining an accurate and easy-to-use ergonomic assessment model stands as an important step in providing feedback for postural correction to improve operator health and comfort. In order to enable efficient computation, previously proposed automated ergonomic assessment and correction tools make approximations or simplifications to gold-standard assessment tools used by ergonomists in practice.
In order to retain assessment quality, while improving computational considerations, we introduce DULA, a differentiable and continuous ergonomics model learned to replicate the popular and scientifically validated RULA assessment. We show that DULA provides assessment comparable to RULA while providing computational benefits. We highlight DULA's strength in a demonstration of gradient-based postural optimization for a simulated teleoperation task.
\end{abstract}

\IEEEpeerreviewmaketitle

\section{Introduction}
\emph{Autonomous Postural Optimization} has received substantial attention in research with the new technologies around humans such as collaborative robots~\cite{peternel2018robot}, smart personal trainers~\cite{chen2020pose}, and VR systems~\cite{hoang2016onebody}. In these systems, the interacting agent should consider human comfort and ergonomics in its behaviour and motion planning \cite{yazdani2021posture}. For example, in a physical human-robot interaction~(pHRI) application such as co-manipulation, one of the objectives in motion planning of the collaborative robot must be satisfying ergonomic safety for the human.


Developing a model of human comfort lies at the heart of effective postural optimization. pHRI researchers have proposed several computational models for assessing ergonomics and human comfort in terms of peripersonal space~\cite{chen2018planning}, muscle fatigue~\cite{peternel2017towards}, and joint overloading~\cite{kim2017anticipatory}. In contrast  ergonomists have provided simpler models which are easier for human experts to calculate by hand and as such are more common in practice. These models include the NASA TLX~\cite{hart1988development}, RULA~\cite{McAtamney1993rula}, REBA~\cite{hignett2000rapid}, strain index~\cite{steven1995strain} and ACGIH TLV~\cite{Kapellusch2014strain}. Importantly these models  are supported by extensive human subject studies that validate their effectiveness on reducing ergonomic risk factors~\cite{ismail2010evaluation, khodabakhshi2014ergonomic}.

Among all risk assessment tools, RULA and REBA depend most on human posture and provide quantitative scores, making them good choices for postural optimization applications. However, the discrete scores and the presence of plateaus in RULA and REBA~\cite{busch2017postural} create challenges when using them in gradient-based postural optimization. Based on our experience, using the risk assessment models directly in gradient-free optimization is time-expensive and the plateaus often prevent progress toward the global optimal solution in postural optimization. Thus, researchers in pHRI often use approximations of ergonomic assessment models in gradient-based postural optimizations; quadratic approximations~\cite{rahal2020caring,busch2017postural,busch2018planning} being the standard approach in the literature. However, these approximations deviate far from the scientifically validated assessments, causing doubt that they can reliably provide the same level of ergonomic  benefit. 

\begin{figure}[t]
    \vspace{2mm}
  \includegraphics[width=8.5cm]{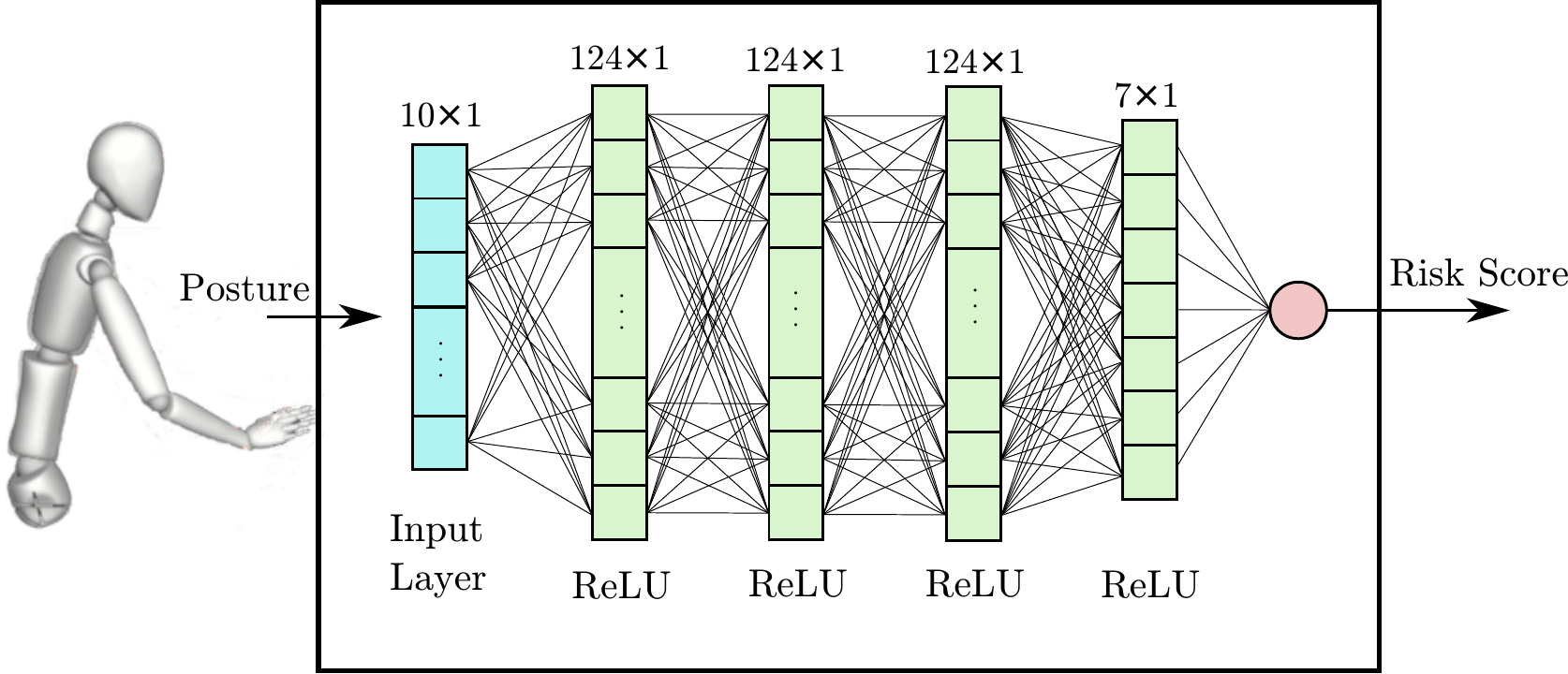}
  \caption{\small The structure of the DULA neural network.}
  \label{fig:network}
  \vspace{-18pt}
\end{figure}

To overcome these issues, we introduce the \textit{Differentiable Upper Limb Assessment}~(DULA), a differentiable and continuous risk assessment model that is learned using a neural network to replicate the popular RULA survey tool (Fig.~\ref{fig:network}). Instead of discrete scores from 1 to 7, DULA reports the risk score as a continuous real number from 1 to 7. Furthermore, it provides the gradient of the risk with respect to each joint enabling efficient use in optimization. We compare the prediction of risk scores of DULA with RULA and provide an open-source package demonstrating how to use DULA in a gradient-based postural optimization in a simulated teleoperation.

\section{Related Work}
\label{sec:related_work}
Although human postural optimization has received significant attention in pHRI, only a few studies~\cite{rahal2020caring,peternel2020human} investigate postural improvement in teleoperation. Table~\ref{table:literature} summarizes the relevant literature for postural optimization in both areas.

\begin{table*}[t!]
\centering
\begin{tabular}{|l|l|l|l|l|l|}
\hline
\multicolumn{2}{|c|}{Ergonomics Model}  & Analytical Models  & Learned Models & \multicolumn{2}{c|}{Risk Assessment Tools}\\ \cline{1-6}
\multicolumn{2}{|c|}{Approximation Method} & & & Quadratic & No Approximation  \\ \cline{1-6} \cline{1-6}
\multirow{6}{*}{\rotatebox[origin=c]{90}{Optimization}} & \multirow{5}{*}{gradient-based} &  \textcolor{grey3}{$\medblacksquare$} Peternel 2018 \cite{peternel2018robot} & & $\meddiamond$ Rahal 2020 \cite{rahal2020caring}  &\\ 
                              & & \textcolor{grey2}{$\medblacksquare$} $\medblacksquare$ Peternel 2017 \cite{peternel2017towards} &  & \textcolor{grey2}{$\medblacksquare$} Busch 2017 \cite{busch2017postural} &   \\
                              &                                   & $\medsquare$ Chen 2018 \cite{chen2018planning} & & \textcolor{grey2}{$\medblacksquare$} Busch 2018 \cite{busch2018planning}  & \\
                              &  & $\medblackdiamond$ Peternel 2020 \cite{peternel2020human} & & &  \\
                              &  & $\medblacksquare$ Kim 2017 \cite{kim2017anticipatory} & & &  \\  \cline{2-6}  
                              & gradient-free &                    & $\medsquare$ Marin 2018 \cite{marin2018optimizing} &              & \textcolor{grey3}{$\medblacksquare$} van der Spaa 2020 \cite{van2020predicting} \\   \hline
\multicolumn{2}{|c|}{Non-optimization} &                                   &                    &                &  $\medsquare$ Shafti et al. \cite{shafti2019real}                \\ \hline                              
\multirow{2}{*}{{Legend}} & \multicolumn{5}{l|}{$\medsquare$ physical HRI: Assistive Holding, \textcolor{grey2}{$\medblacksquare$} physical HRI: Handover, \textcolor{grey3}{$\medblacksquare$} physical HRI: Co-Manipulation} \\
&\multicolumn{5}{l|}{{$\medblackdiamond$} Teleop: Goal-Constrained with Repositioning Postural Correction, $\meddiamond$ Teleop: Goal-Constrained with Online Postural Correction} \\
\hline
\end{tabular}
\caption{\small State of the art of postural optimization in pHRI and teleoperation.}
\label{table:literature}
\vspace{-0.2in}
\end{table*}

Researchers have examined ergonomics and postural optimization in three different types of physical human-robot interaction scenarios: (1) \textit{Assistive Holding} (e.g.~\cite{marin2018optimizing,shafti2019real,chen2018planning}), (2) \textit{object handover} (e.g.~\cite{busch2017postural,busch2018planning,peternel2017towards}), and (3) \textit{Co-Manipulation} (e.g.~\cite{van2020predicting,peternel2018robot,peternel2017towards,kim2017anticipatory}). In the first two tasks the postural optimization provides optimized posture for the human and joint configurations for the robot while maintaining contact with the object at the interaction interface. For the case of co-manipulation, the postural optimization outputs a trajectory of optimal postures for the human and an optimal joint-space trajectory for the robot to perform the co-manipulation task. 


Ergonomic assessment models provide the primary cost in postural optimization objectives. pHRI researchers have proposed many computational models to assess ergonomics and human comfort of users including peripersonal space~\cite{chen2018planning}, muscle fatigue~\cite{peternel2017towards}, and joint overloading~\cite{kim2017anticipatory}. To make the optimization simpler, Marin et al. suggested the idea of a contextual ergonomics model which is a set of Gaussian process models including joint angles, moments, reaction, load and muscle activation, trained with the musculoskeletal simulation task contexts~\cite{marin2018optimizing}. Using Gaussian process models enables search in a 2D latent space while their cost function is defined in the high-dimensional musculoskeletal space. 

Some literature use approximations of the ergonomics risk assessment tools. Busch et al. proposed a differentiable surrogate of the REBA score by fitting a weighted combination of quadratic functions plus a constant for the task payload~\cite{busch2018planning}. Rahal et al. suggested a quadratic approximation for RULA which is the summation of the quadratic norm of the deviation from the human neutral posture for shoulder, elbow and wrist joints angles~\cite{rahal2020caring}. Their approximation conceptually agrees with the qualitative idea behind RULA in which the risk score goes higher when the human deviates more from the neutral posture. 
Van der Spaa et al. provide the only study of directly using risk assessment tools in derivative-free postural optimization~\cite{van2020predicting}. They add the REBA score to the transition cost function in an A* optimization for task and motion planning of a robot in a co-manipulation task.

\begin{figure}[t]
    \hspace{-2mm}
\center
  \includegraphics[width=8cm]{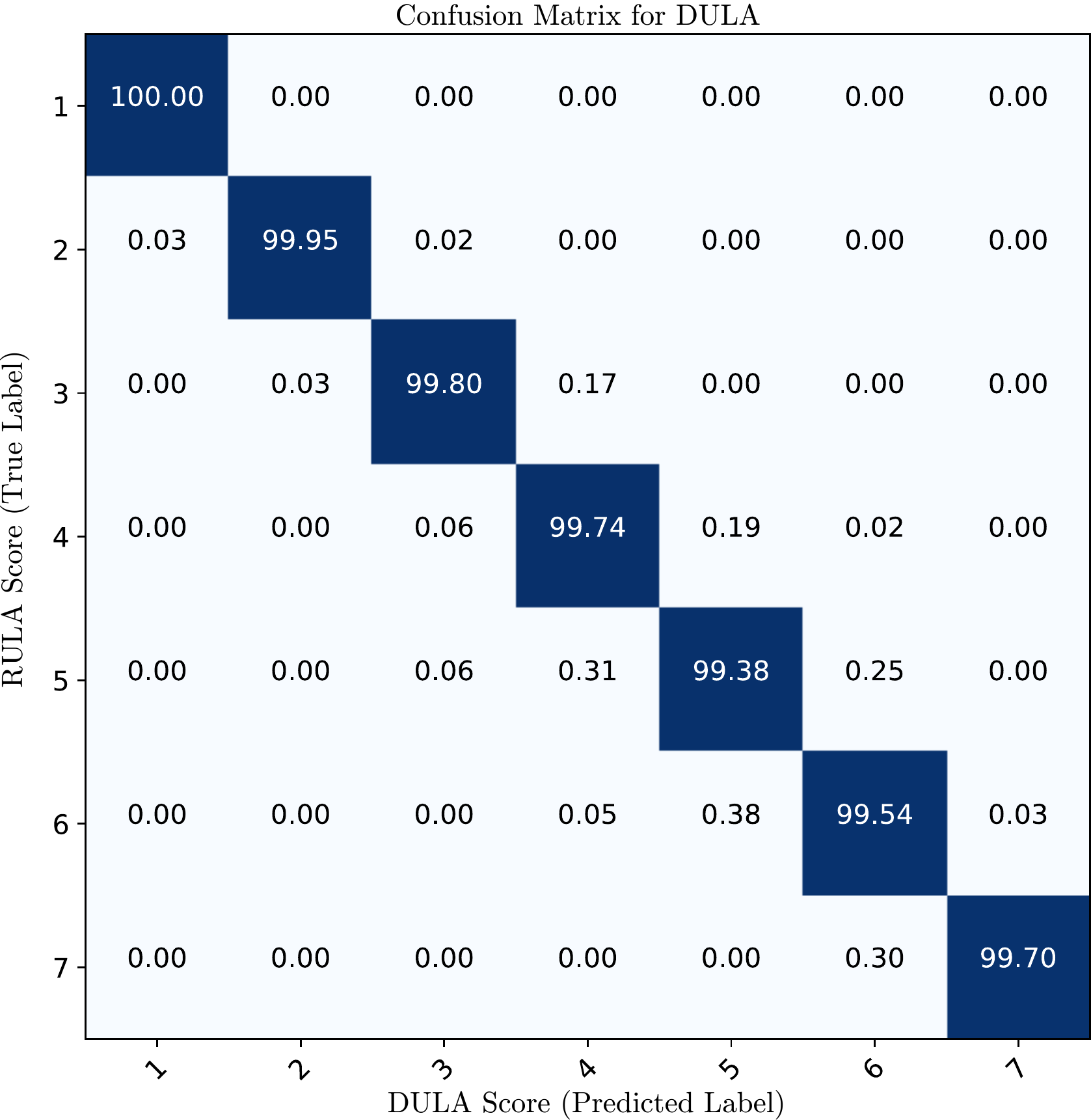}
  \caption{\small Confusion matrix (accuracy \%) for DULA vs RULA.}
  \label{fig:confusion}
  \vspace*{-6.5mm}
\end{figure}

\section{Differentiable Human Ergonomics Model}
To build a differentiable RULA, we developed a dataset of 7.5 million upper body postures of a human model consisting of 3 joints in the torso and 7 joints in the arm. We additionally define task parameters based on the RULA worksheet---the frequency of the arm and body motions; type and maximum load on the arm and body; neck angle; and whether any legs, feet, or arms are supported. As the human range of motion is pose dependant, to ensure the validity of the postures, we used the learned pose-dependant model of posture validity provided by~\cite{jiang2018data}. We developed a script for automatic RULA assessment based on the posture and tasks parameters and verified it with several ergonomists. We used this script to label the posture dataset. Since postures with labels 1, 2, 6, and 7 are not frequent in the full range of human motion, we balanced the dataset by forcing the data generation scripts to generate enough data points with those labels. We split the dataset into 80\% training and 20\% testing sets. 

Moreover, to learn a continuous and differentiable function for RULA, we designed a fully-connected regression-based neural network. While RULA provides discrete integer scores from 1 to 7, we choose to predict continuous labels. If we had instead performed multi-class classification based on the discrete labels the resulting model would be less useful in optimization as there is no natural choice of smooth objective to minimize or constrain ergonomic cost. This would negate our desire for a computationally useful model. Hence, we perform regression to the multi-class labels.  

The structure of the neural network is shown in Fig.~\ref{fig:network}. It includes 4 hidden layers with ReLU activation function. We found that a network with 124 units for the first three layers and 7 units for the last hidden layer worked best. We train the network using the standard mean squared error loss function for 2000 epochs using a learning rate of 0.001. We used 5-fold cross-validation to find the optimal network parameters.
 This results in a model with 99.73\% accuracy. We round our continuous output to the nearest integer when reporting accuracy. Figure~\ref{fig:confusion} shows the confusion matrix for the learned DULA model. The lowest diagonal element is 99.38\% which shows the high accuracy of the learned model across all ranges.

\section{Postural Optimization Using DULA}
We use the learned DULA model in a gradient-based postural optimization for a simple teleoperation scenario in which a human interacts with a leader robot to remotely control a follower robot. As the human performs the task, the intelligent teleoperation system estimates the human posture using e.g. marker-based or markerless posture estimation, or directly from the leader robot using the method proposed in~\cite{yazdani2020leader}. Then, it performs risk assessment to obtain the ergonomics risk score using the standard RULA model. The postural optimization algorithm uses DULA to find the optimal posture that results in the same hand pose at the interaction point between the human and the leader robot, while having the minimum risk of injuries and then provides the user with the online optimal postural correction to move towards while completing the task.

The human operator can correct the posture while performing the task without pausing. This approach is beneficial for goal-constrained teleoperation tasks such as pick-and-place in which the goal position for placing the object is defined, but, the path toward the goal point is not constrained.
\begin{figure}[t]
\center
\includegraphics[width =8.8cm]{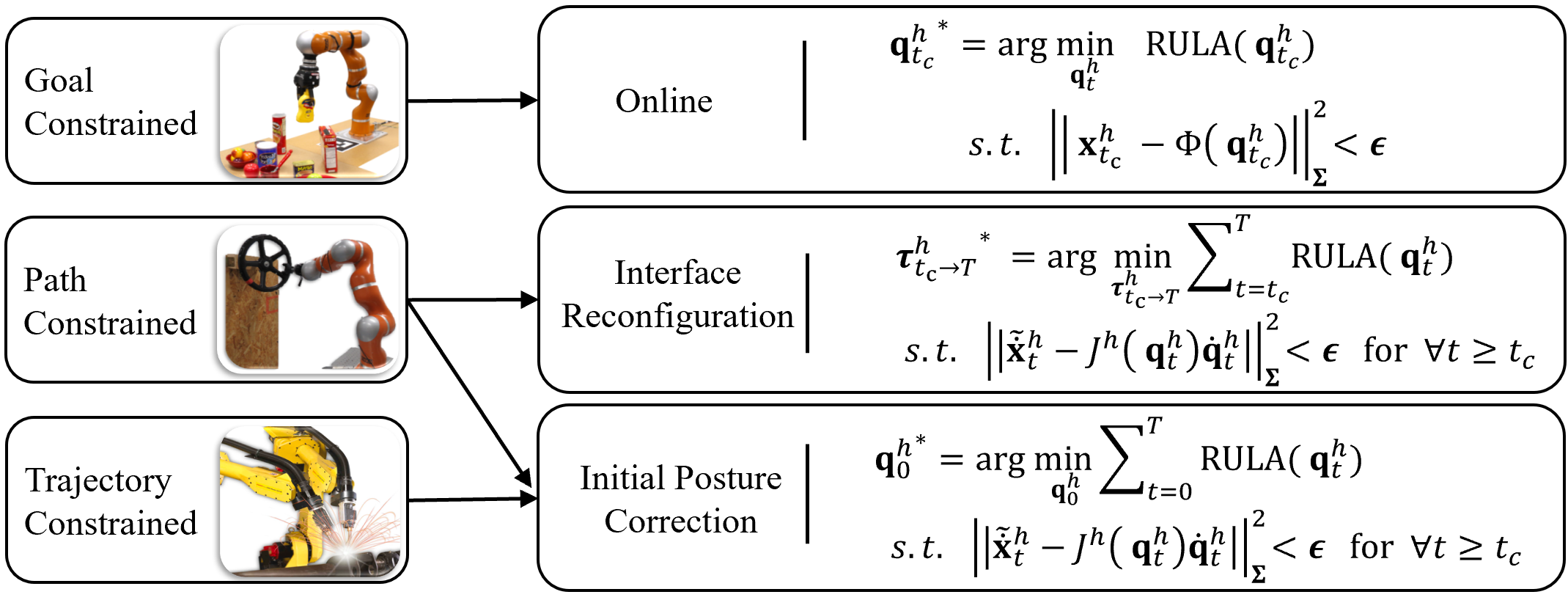}
\vspace{-0.6cm}
\caption{\small Postural optimization categories in teleoperation.}
\label{fig:teleopt}
\vspace{-0.4cm}
\end{figure}

 \begin{figure}[t!]
\center
\includegraphics[width = 7cm]{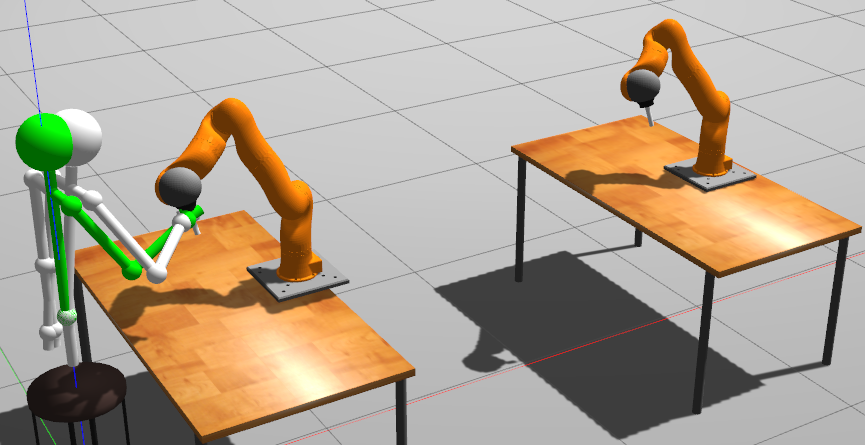}
\caption{\small Teleoperation simulation environment in ROS. The green skeleton visualizes the suggested optimal posture and the white skeleton shows the simulated human.}
\label{fig:ros-sim}
\vspace{-0.6cm}
\end{figure}

We define online postural optimization for teleoperation as:
\begin{flalign}
\mathbf{q}^*_t = \text{arg}\underset{\mathbf{q}_t}{\text{min}}&\quad
\mathrm{DULA}(\mathbf{q}_t) \label{eq:opt_online}\\
\text{s.t.}& \quad ||\mathbf{x}_t-\Phi(\mathbf{q}_t)||^2_{\Sigma}<\epsilon \nonumber\\
            & \mathbf{q}_t \in \text{Range of Motion} \nonumber 
\end{flalign}
where ${\mathbf{q}}^*_t$ and ${\mathbf{q}}_t$ are optimal posture and posture at time $t$, respectively, $\mathbf{x}_t$ is the observed pose of the hand measured by the leader robot, $\Phi$ is the forward kinematics of the human, and $\Sigma$ is the weight vector for position and orientation elements.

It is important to note that the optimal posture 
${\mathbf{q}}^*_t$  from Eq.~\eqref{eq:opt_online} for each time step is then suggested to the human to move towards. The human can refuse or accept, and try to apply it as much as possible while completing the task.

\begin{figure}[t]
\center
\includegraphics[width = 7.5cm]{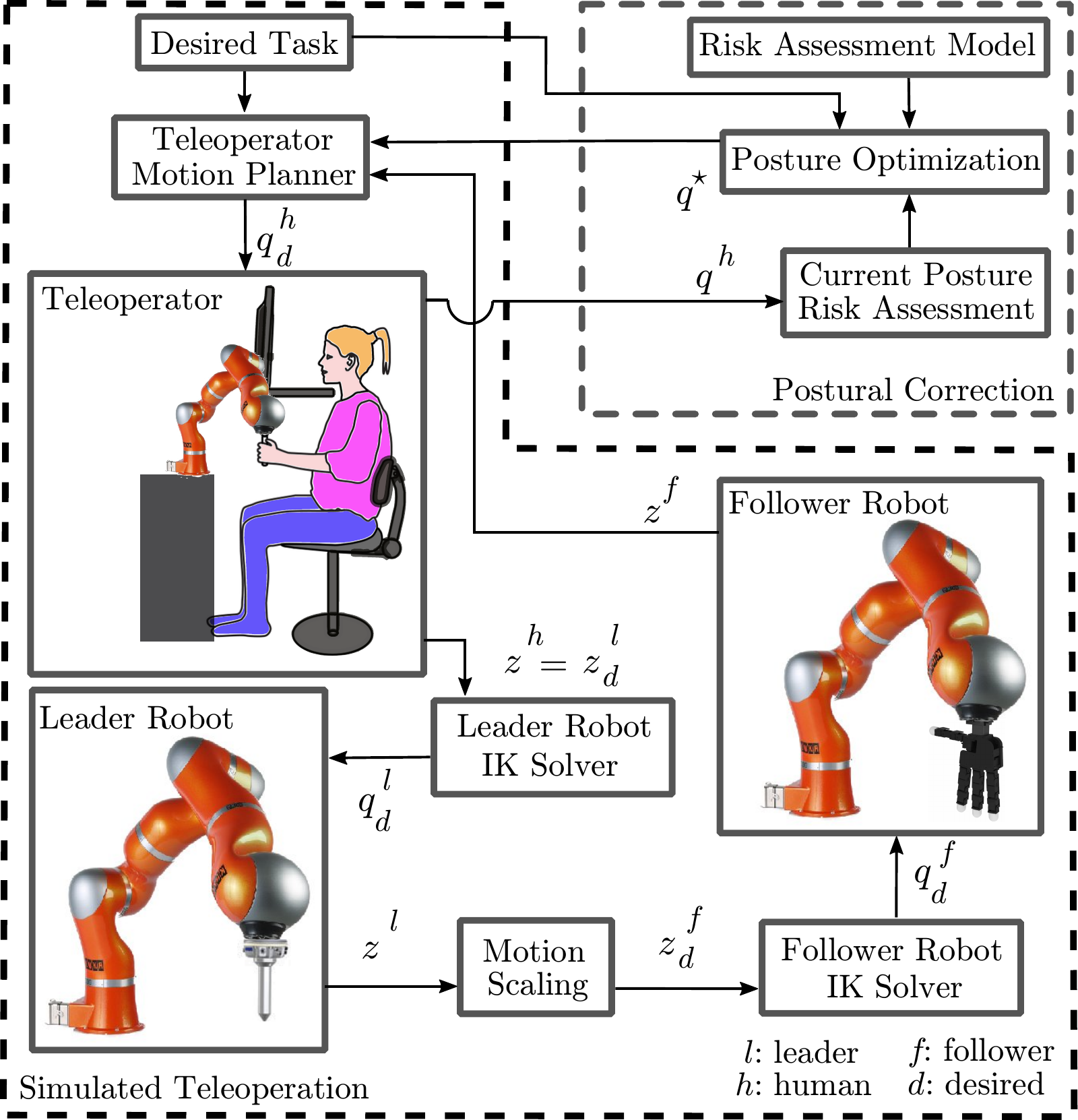}
\vspace{-0.2cm}
\caption{\small Overview of teleoperation simulation.}
\label{fig:framework}
\vspace{-0.6cm}
\end{figure}

In addition to goal-constrained teleoperation, we also defined and formulated the use of DULA in postural optimization for path-constrained teleoperation~(e.g. turning a valve where the path to follow is constrained based on the diameter of the valve) and trajectory-constrained teleoperation~(e.g. arc welding where the operator should follow a velocity profile in addition to the welding path). Figure~\ref{fig:teleopt} summarizes our formulation for different types of teleoperation tasks and their corresponding postural optimization approaches. We focus on online postural correction in this paper; analyzing the other problem formulations is ongoing work.

\begin{figure}[t!]
\vspace{-1mm}
\centering
\begin{subfigure}{0.5\textwidth}
\centering
\includegraphics[width = .72\linewidth]{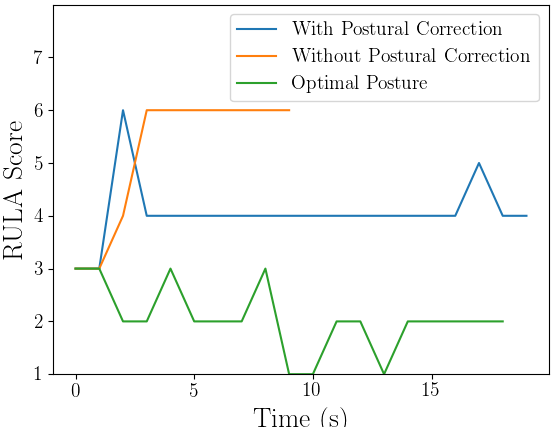}
\vspace{-2mm}
\subsection{\small Gradient-free postural optimization}
\end{subfigure}
\begin{subfigure}{0.5\textwidth}
\centering
\includegraphics[width = .72\linewidth]{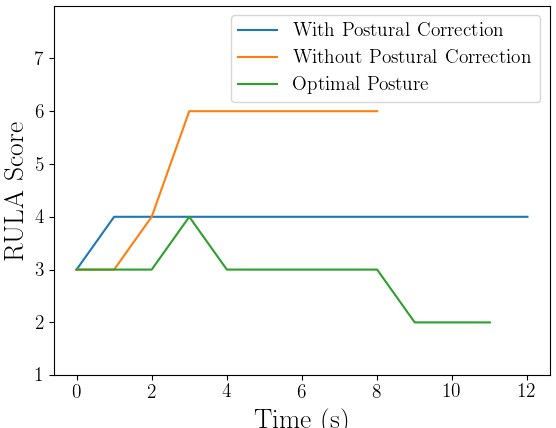}
\vspace{-2mm}
\subsection{\small Gradient-based postural optimization}
\end{subfigure}
\caption{\small Comparison of gradient-free and gradient-based postural correction on a teleoperation task. Lower scores are better.}
\label{fig:result1}
\vspace{-0.7cm}
\end{figure}

We used a sequential quadratic programming~(SQP)~\cite{nocedal2006numerical} solver from SciPy, bounded on the range of motion, to solve the nonlinear optimization in Eq.~\eqref{eq:opt_online}, and calculated DULA gradients using automatic differentiation in PyTorch.

\section{Simulated Teleoperation Environment}
We developed an open-source simulator for postural correction in teleoperation using ROS. It includes a human seated on a stool, and two 7-DOF KUKA LWR-4 robots as leader and follower robots as shown in Fig.~\ref{fig:ros-sim}. We model our simulated human operator to behave like a human in two ways: (1) physically controlling the teleoperation task and (2)~accepting or rejecting the recommended postural corrections. 

We model this as an optimal motion planning framework with re-planning that finds a human joint trajectory that controls the follower robot for the desired task while moving toward the optimal ergonomic posture:
\begin{flalign}
{\mathbf{\tau}^h_{t\rightarrow H}}^* = \text{arg}\underset{\mathbf{\tau}^h_{t\rightarrow H}}{\text{min}}\sum_{t=t}^H||\mathbf{x}^f_g - \mathbf{x}^f_t||^2_{\Sigma}  + \alpha||\mathbf{q}^h_t - {\mathbf{q}^h}^*_t||_2^2
\label{eq:human_motion_with_correction}
\end{flalign}
where $\mathbf{\tau}^f_{t\rightarrow H}$ is the trajectory of human posture from time $t$ to the time of the horizon $H$, $\mathbf{x}^f_g$ is the goal pose of the follower robot, and $^f\mathbf{x}_t$ is the pose of the follower robot at time $t$. $\mathbf{q}_t$ and $\mathbf{q}^*_t$ are the current posture and optimal posture of the human from the postural optimization at time $t$, respectively. Here, $0\leq \alpha \leq 1 $ is a scalar number that models the postural correction acceptance and the effort of the human operator towards applying the postural correction. Details of the motion planning for the human and the robots are presented in Fig.~\ref{fig:framework}. We note that this model is likely a simplification of true human teleoperation behavior. We do not advocate for its use over human subject studies. Instead, we propose it as a useful tool for systematically exploring new algorithms in human safety assessment and improvement.

As a comparison to DULA, we directly use the standard RULA in a gradient-free postural optimization using the cross-entropy method~\cite{de2005tutorial}. Figure~\ref{fig:result1}(A) shows the postural correction using this approach in a simple teleoperation task. It presents the RULA scores for calculated optimal posture~(green), human posture without postural correction~(orange), and the human posture after applying the postural correction~(blue) according to the human control and acceptance model during the task. The plot shows that the risk is reduced after the simulated human applies the suggested optimal postural correction. Also the task completion time increases without a significant decrease in risk.

Figure~\ref{fig:result1}(B) shows the postural correction for the same task using gradient-based optimization with DULA. We can see that gradient-based approach provides a smoother motion with respect to the RULA risk score and avoids going through postures with risk higher than 4. It also results in shorter task completion time than the gradient-free approach.

\begin{figure}[t]
\center
\includegraphics[width = 8cm]{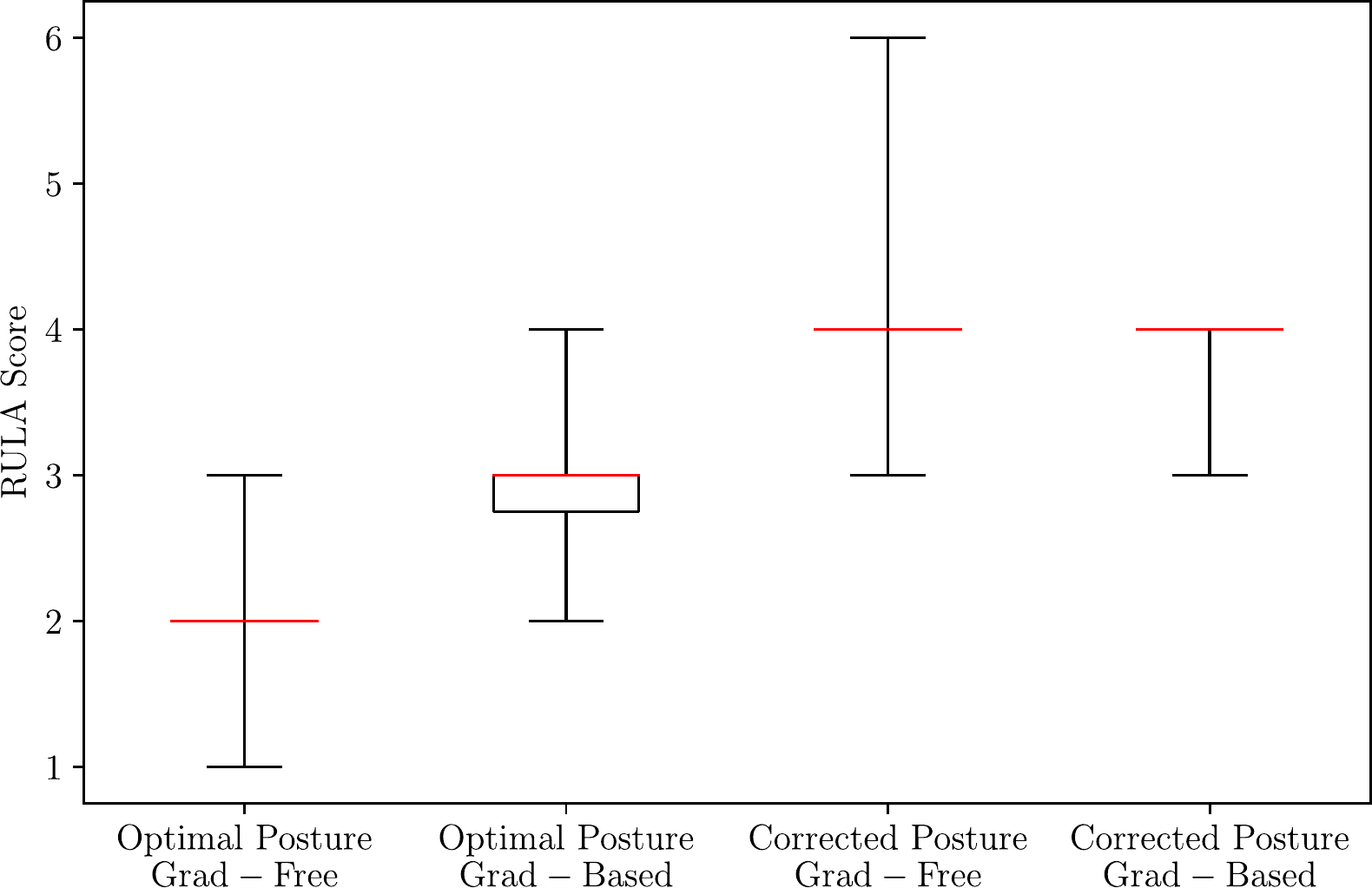}
\caption{\small comparing the optimal posture and corrected posture between gradient-free optimization using RULA and gradient-based optimization using DULA. Lower scores are better. An iteration of the gradient-free method takes 2.7 minutes, where the gradient-based method takes only tenths of a second.}
\label{fig:comparison}
\vspace{-0.6cm}
\end{figure}

Figure~\ref{fig:comparison} provides more information on comparing gradient-free and gradient-based postural optimization. It shows that the median risk scores of target optimal postures from the gradient-free approach is lower. However, the postures of the simulated human are more comfortable after applying the optimal posture calculated from the gradient-based optimization. We believe it is due to the smoother optimal postures that has been suggested to the simulated human from the gradient-based method. Moreover, the gradient-based approach is much faster. Each iteration of the gradient-free approach using 10000 samples takes 2.7 minutes to solve, while the gradient-based approach takes only tenths of a second.

\section{Conclusion and Future Directions}
We introduced DULA, a differentiable and continuous ergonomics model to assess human upper body posture. DULA learned to replicate the non-differentiable RULA using a neural network. The proposed model is 99.73\% accurate and computationally designed for effecient use in postural optimization for pHRI and other related applications. We also introduced a framework for postural optimization in teleportation using DULA and presented a demo task. The results reveal postural optimization using DULA lowers the risk score for goal-constrained teleop. The trained DULA model, data, algorithm, and demo are available as an open-source package\footnote{\url{https://sites.google.com/view/differentiable-ergonomics}}.

There are several directions for future work. As a fist step, we plan to follow the same procedure for REBA a whole body assessment tool. Additionally, we intend to conduct a human subject study to evaluate our postural optimization and correction approach. We wish to compare different means of feedback for the posture correction to the human operator including visual, auditory, and haptic feedback.
\section{Aknowledment}
\label{sec:conclusion}
This work is supported by DARPA under grant N66001-19-2-4035.

\bibliographystyle{plainnat}
\bibliography{references}

\end{document}